\documentclass{article}

\usepackage[preprint]{neurips_2025}

\usepackage[utf8]{inputenc} % allow utf-8 input
\usepackage[T1]{fontenc}    % use 8-bit T1 fonts

\usepackage{hyperref}       % hyperlinks
\usepackage{url}            % simple URL typesetting

\usepackage{graphicx}
\usepackage{caption}
\usepackage{subcaption}
\usepackage{float}
\usepackage[dvipsnames,table,usenames]{xcolor}

\usepackage{amsmath}
\usepackage{amsfonts}
\usepackage{nicefrac}
\usepackage{siunitx}
\usepackage{cleveref}

\usepackage{booktabs}
\usepackage{multirow}
\usepackage{array}

\usepackage{microtype}
\usepackage{gensymb}
\usepackage{bbding}

\title{Look Before Acting: Enhancing Vision Foundation Representations for Vision-Language-Action Models}

\author{
Yulin Luo\textsuperscript{\rm 1$^{*}$}, 
Hao Chen\textsuperscript{\rm 3$^{*,\dagger}$}, Zhuangzhe Wu\textsuperscript{\rm 1$^{*}$}, 
Bowen Sui\textsuperscript{\rm 1$^{*}$}, 
Jiaming Liu\textsuperscript{\rm 1$^{*,\dagger}$}, \\ 
\textbf{Chenyang Gu}\textsuperscript{\rm 1}, 
\textbf{Zhuoyang Liu}\textsuperscript{\rm 1},
\textbf{Qiuxuan Feng}\textsuperscript{\rm 1}, 
\textbf{Jiale Yu}\textsuperscript{\rm 1}, 
\textbf{Shuo Gu}\textsuperscript{\rm 2}, 
\textbf{Peng Jia}\textsuperscript{\rm 2}, \\  
\textbf{Pheng-Ann Heng}\textsuperscript{\rm 3}, 
\textbf{Shanghang Zhang}\textsuperscript{\rm 1,\Envelope} \\[0.5em]
\textsuperscript{\rm 1}State Key Laboratory of Multimedia Information Processing, School of Computer Science, \\
Peking University; 
\textsuperscript{\rm 2}Simplexity Robotics; 
\textsuperscript{\rm 3}The Chinese University of Hong Kong\\[0.5em]
$^{*}$ Equal contribution, $^{\dagger}$ Project lead, \Envelope\ Corresponding author\\[0.5em]
\textbf{Project page}: \href{https://deepvision-vla.github.io/}{https://deepvision-vla.github.io/}
}

\begin{document}

\maketitle
\begin{abstract}
Vision-Language-Action (VLA) models have recently emerged as a promising paradigm for robotic manipulation, in which reliable action prediction critically depends on accurately interpreting and integrating visual observations conditioned on language instructions. Although recent works have sought to enhance the visual capabilities of VLA models, most approaches treat the LLM backbone as a black box, providing limited insight into how visual information is grounded into action generation. Therefore, we perform a systematic analysis of multiple VLA models across different action-generation paradigms and observe that sensitivity to visual tokens progressively decreases in deeper layers during action generation. Motivated by this observation, we propose \textbf{DeepVision-VLA}, built on a \textbf{Vision-Language Mixture-of-Transformers (VL-MoT)} framework. This framework enables shared attention between the vision foundation model and the VLA backbone, injecting multi-level visual features from the vision expert into deeper layers of the VLA backbone to enhance visual representations for precise and complex manipulation. In addition, we introduce \textbf{Action-Guided Visual Pruning (AGVP)}, which leverages shallow-layer attention to prune irrelevant visual tokens while preserving task-relevant ones, reinforcing critical visual cues for manipulation with minimal computational overhead. DeepVision-VLA outperforms prior state-of-the-art methods by 9.0\% and 7.5\% on simulated and real-world tasks, respectively, providing new insights for the design of visually enhanced VLA models.

\end{abstract}   
\section{Introduction}
Driven by training on massive, internet-scale multimodal corpora \cite{li2024llava1,li2024omnicorpus,li2024llava,schuhmann2022laion,awadalla2024mint}, recent advances in Vision-Language Models (VLMs) have demonstrated exceptional proficiency in perception, reasoning, and instruction following \cite{lu2024deepseek,chen2025janus,bai2025qwen3,karamcheti2024prismatic,beyer2024paligemma, luo2025robobench}. Capitalizing on these strengths, Vision-Language-Action (VLA) models extend these capabilities to robotics by directly mapping multimodal observations and natural language instructions to robot actions. Powered by billions of parameters and large-scale robot pre-training datasets \cite{o2024open,khazatsky2024droid,wu2024robomind}, VLAs have exhibited remarkable potential for learning generalizable manipulation skills across diverse scenarios.

The robust control and generalization of VLA models are fundamentally contingent upon the precise interpretation and integration of visual observations \cite{zhang2026vlm4vla}. Consequently, contemporary research has increasingly focused on fortifying the visual understanding and reasoning of VLAs. 
Existing approaches enhance the visual capabilities of VLA models from four main perspectives: 
\begin{figure}
    \centering
    \includegraphics[width=\textwidth]{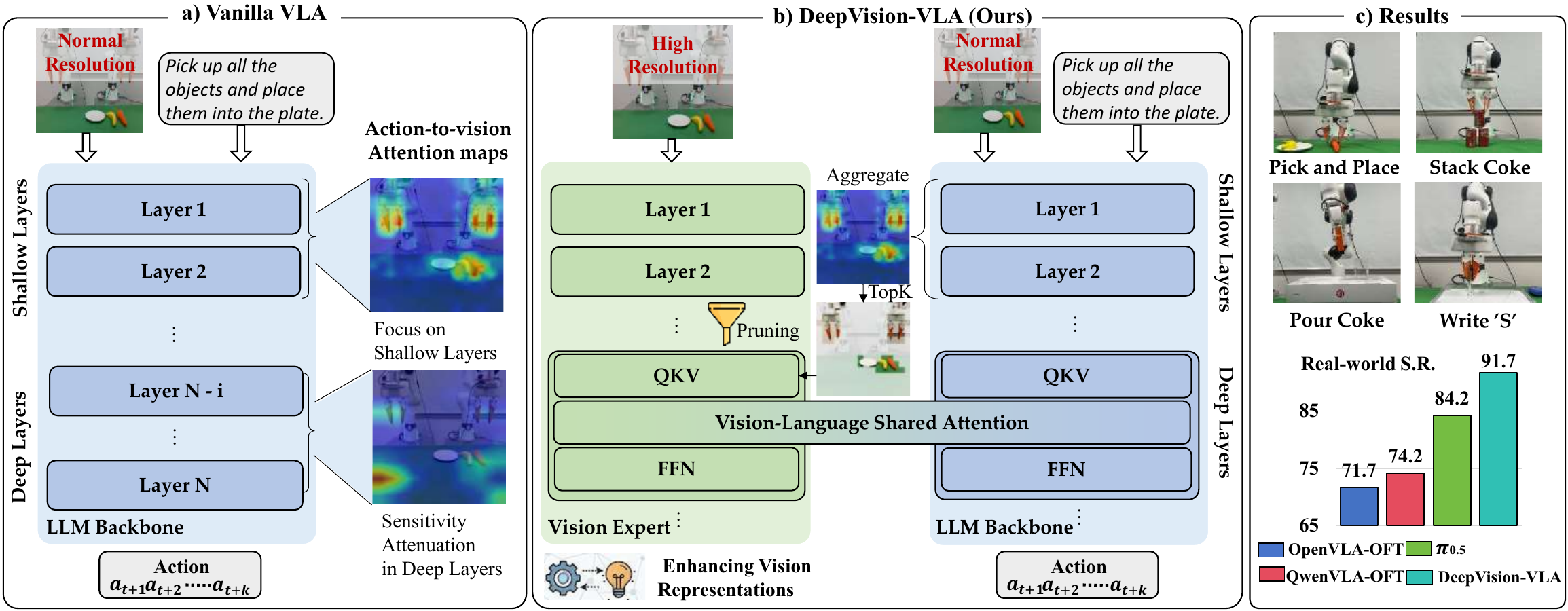}
    \caption{
    \textbf{Overview.} (a) In vanilla VLA models, attention to task-relevant visual tokens gradually weakens in deeper layers, leading to reduced sensitivity to visual information. (b) To address this issue, DeepVision-VLA introduces a Vision-Language Mixture-of-Transformers framework that injects multi-level visual features from a vision foundation model into deeper layers of the VLA backbone, strengthening visual representations for precise and complex manipulation. (c) Built upon the custom QwenVLA-OFT, DeepVision-VLA significantly outperforms the baseline by a large margin and achieves superior performance across several real-world manipulation tasks.
    }
    \vspace{-1cm}
    \label{fig:teaser}
\end{figure}
(1) introducing visual prompts to improve the model’s understanding of scenes and manipulated objects~\cite{gu2023rt, sundaresan2024rt, li2025object}; (2) designing auxiliary visual objectives to encourage the model to focus on task-critical entities~\cite{kachaev2025don,song2025reconvla}; (3) incorporating additional visual modalities to provide complementary information \cite{yuan2025depthvla,zhen20243d,li2026pointvla,sundaresan2024rt}; and (4) predicting future states to strengthen the model’s ability to model the physical world~\cite{zhao2025cot, liu2025mla, liu2026last}.
Despite these advancements, existing paradigms typically treat the underlying Large Language Model (LLM) backbone in VLA models as a monolithic "black box", offering little insight into how visual information is propagated and utilized. 

In this work, we move beyond treating the LLM backbone in VLA models as an opaque module and instead investigate how visual information is processed across its internal layers. Since the LLM is composed of stacked Transformer layers, a layer-wise analysis offers a natural and granular lens for understanding multimodal integration.
We conduct a systematic investigation of several representative VLA architectures and action generation paradigms in two complementary stages. First, we qualitatively analyze action-to-visual attention maps and observe that while early layers maintain grounded attention on task-relevant objects, deeper layers often fail to focus effectively on these regions.
To quantify the impact of this phenomenon on action generation, we introduce a layer-wise visual token dropout strategy. The results align with our qualitative findings: in deeper layers, action accuracy becomes increasingly insensitive to the masking of task-relevant visual regions, whereas shallow layers exhibit the opposite trend.
We attribute this pattern to the prevalent serial architecture of current VLA models, where visual information is injected only at the first LLM layer and gradually attenuates as it propagates through Transformer layers.

Motivated by these findings, as illustrated in Figure~\ref{fig:teaser}, we introduce \textbf{Vision-Language Mixture-of-Transformers (VL-MoT)}, a novel VLA framework designed to enhance the sensitivity of deeper layers to task-relevant visual regions, thereby improving action prediction. In addition to the original visual encoder, our framework incorporates a dedicated \textbf{Vision Expert} (DINOv3, 0.8B) whose representations are fused with the VLA backbone via a MoT mechanism. Building on our observation that shallow layers naturally maintain effective visual grounding, we adopt an integration scheme that selectively couples the Vision Expert only with deeper VLA layers.
Specifically, we extract multi-level visual features by sampling from the last few Transformer layers of the Vision Expert. This strategy empirically outperforms alternative integration approaches, such as sampling from the early layers or uniformly across all layers. A plausible explanation is that the later layers of the Vision Expert capture higher-level, semantically rich representations that are more invariant and object-centric, making them more compatible with task-relevant, action-conditioned features in the VLA model.
Through this targeted integration, the Vision Expert collaborates with deeper VLA layers to reinforce action generation where the model is most susceptible to visual degradation, enabling more reliable and visually grounded robotic control.

However, naively integrating the full feature maps from the Vision Expert into the VLA backbone may introduce significant redundancy and irrelevant background information, potentially diluting task-critical signals. To address this, we propose an \textbf{Action-Guided Visual Pruning (AGVP)} strategy that refines information flow within our framework. Specifically, we leverage the robust grounding capabilities of shallow VLA layers to compute a saliency map by averaging attention weights from action tokens to visual tokens. This map identifies the most task-relevant regions, which we then use to prune the Query, Key, and Value across the Vision Expert's Transformer layers before integrating them into the deep VLA layers. This targeted pruning not only mitigates visual redundancy but also enables the Vision Expert to process higher-resolution inputs with minimal computational overhead. Our empirical results show that the increased visual granularity, focusing precisely on action-critical entities, leads to more stable manipulation.

We instantiate our framework as \textbf{DeepVision-VLA}, built upon a custom baseline, QwenVLA-OFT. This baseline leverages a Qwen3-VL backbone (4B) and adopts parallel action decoding with L1 regression output \cite{kim2025fine}.
We systematically evaluate DeepVision-VLA across ten simulated tasks in RLBench \cite{james2020rlbench} and four complex dual-arm real-world manipulation tasks. DeepVision-VLA achieves state-of-the-art (SOTA) performance, outperforming prior VLA methods by 9.0\% in simulated settings and 7.5\% in real-world settings.
Our contributions are summarized as follows:

\begin{itemize}
\item We systematically analyze visual information utilized in current VLA models and identify a phenomenon where deeper LLM backbones become insensitive to task-relevant visual regions.
    
\item We propose Vision-Language Mixture-of-Transformers (VL-MoT), a novel framework that improves action prediction by injecting multi-level visual features from a Vision Expert into deep VLA layers.
    
\item We introduce Action-Guided Visual Pruning, a strategy that filters redundant visual information from the Vision Expert, providing deeper VLA with task-relevant visual signals.
    
\item DeepVision-VLA establishes SOTA results in both simulation and real-world settings, demonstrating the effectiveness of our framework and providing more insights for the design of visually enhanced VLA models.
\end{itemize}

\section{Related Work}
\textbf{Vision-Language-Action (VLA) Models.}
VLA models~\cite{liu2024rdt, wen2025diffusionvla, wen2025tinyvla, liu2025hybridvla, black2024pi_0, intelligence2025pi_, bjorck2025gr00t, belkhale2024rt, kim2024openvla,chen2025fast} are primarily driven by scaling robot demonstration data \cite{wu2024robomind,o2024open,khazatsky2024droid,bu2025agibot} and adapting pretrained vision-language models (VLMs) \cite{bai2025qwen3,karamcheti2024prismatic,lu2024deepseek,chen2025janus} for robotic control. These approaches directly model action sequences from visual observations and language instructions, demonstrating strong scalability and significant potential for generalization. Early work attempted to leverage the autoregressive capabilities of pretrained VLMs to generate robot actions token by token \cite{kim2024openvla,zitkovich2023rt,belkhale2024rt,brohan2022rt}. However, such formulations often suffer from action discontinuities and low execution frequency. Inspired by the success of diffusion policies \cite{ze20243d,chi2025diffusion}, recent efforts have explored diffusion-based \cite{wen2025diffusionvla,li2024cogact,liu2025hybridvla} and flow-based VLA \cite{black2024pi_0,intelligence2025pi_} frameworks, which leverage the strong representation power of VLMs while introducing a dedicated action head to learn smooth and stable continuous action outputs.
To further improve execution efficiency, several works \cite{chen2025fast,cui2025openhelix,bu2024towards} adopt a dual-system design, where a reasoning module is responsible for task planning, while a control module focuses on action generation. In addition, hierarchical architectures~\cite{shi2025hi} have been proposed to better handle high-level and abstract human instructions, typically leveraging an auxiliary VLM to decompose complex instructions into subgoals that guide the VLA for downstream instruction-following action generation.
However, recent studies~\cite{zhang2026vlm4vla, kachaev2025don,song2025reconvla} suggest that when VLA models do not sufficiently develop their visual understanding during training, their action modeling performance can degrade noticeably, which may hinder precise manipulation in dynamic or cluttered environments. These findings suggest that preserving reliable, task-aware visual representations can be crucial for effective action generation~\cite{fei2025libero, yang2026uaor, tang2025uad}.

\textbf{Vision Improvement for VLA.}
As precise action generation critically depends on robust visual understanding and grounding~\cite{zhang2026vlm4vla, song2025reconvla}, a growing body of work has explored strategies to enhance VLA models from a visual perspective.
One line of research augments the VLA input with additional visual cues to facilitate task comprehension, such as overlaying execution trajectories \cite{gu2023rt,li2025object} or highlighting target objects \cite{gu2025manualvlaunifiedvlamodel}, demonstrating that simple prompt engineering can be surprisingly effective.
Another approach introduces auxiliary visual supervision to encourage the model to attend to important image regions, for example, by reconstructing key objects in the image \cite{song2025reconvla} or anchoring the VLA’s visual representations to strong teacher features \cite{kachaev2025don}, thereby improving the reliability of action generation.
Beyond 2D inputs, incorporating richer visual modalities such as depth maps \cite{tur2026recurrent,li2025qdepth,yuan2025depthvla}, 3D point clouds \cite{zhen20243d,li2026pointvla,qu2025spatialvla}, or hand-drawn sketches \cite{sundaresan2024rt} provides complementary spatial and geometric information, enabling the model to better reason about object shapes, distances, and occlusions.
Finally, several methods adopt a reasoning-before-action paradigm \cite{liu2026last,zhao2025cot,cen2025worldvla}, predicting future states or images to strengthen the model’s understanding of physical dynamics, thereby enhancing the accuracy of manipulation.
Despite these advancements, existing paradigms often treat the visual processing within VLA models as a black box, focusing primarily on the input state and the resulting actions while largely overlooking how visual information is internally utilized. In this work, we provide insights into this process and propose DeepVision-VLA, which integrates multi-level features from a visual foundation model into the deeper layers of the VLA via a Vision-Language Mixture-of-Transformers architecture, enhancing attention to task-relevant objects and improving action generation accuracy.

\section{Methods}
In this section, we first introduce the problem formulation and the general architecture of Vision-Language-Action (VLA) models in Sec.~3.1.
Next, Sec.~3.2 analyzes how representative VLA architectures process and utilize visual information, providing key insights into their limitations.
Building on this analysis, we propose the \textbf{Vision-Language Mixture-of-Transformers (VL-MoT)} framework, which improves action prediction performance by injecting multi-level Vision Expert knowledge into the deep layers of VLA models.
Sec.~3.3 presents \textbf{DeepVision-VLA}, a concrete instantiation of this framework, and introduces the \textbf{Action-Guided Vision Pruning (AGVP)} strategy to further enhance the model's focus on task-relevant visual regions. The overall framework is shown in Figure \ref{fig:method}.
Finally, Sec.~3.4 describes the training and inference procedures of DeepVision-VLA.

\subsection{Preliminaries}

\textbf{Problem Formulation.}
We consider a standard imitation learning setting for vision-language robotic manipulation. 
Given a dataset of expert demonstrations $\mathcal{D}=\{(\tau_i, l_i)\}_{i=1}^{N}$, where $l_i$ denotes the language instruction and $\tau_i=\{(o_t, a_t)\}_{t=1}^{T_i}$ represents the corresponding trajectory consisting of visual observations and actions, the goal is to learn a policy $\pi_\theta$ that maps visual observations and task instructions to robot actions. 
At each time step $t$, the policy takes the current observation $o_t$ together with the instruction $l$ as input and predicts robot actions. Depending on the action representation, the policy may either predict the current action $a_t$ or a short horizon of future actions $a_{t:t+H-1}$ under an action chunking formulation. Formally, this can be written as $\pi_\theta(a_t \mid o_{\le t}, l)$ or $\pi_\theta(a_{t:t+H-1} \mid o_{\le t}, l)$. The policy parameters are learned from demonstrations by solving
\begin{equation}
\theta^{*}=\arg\min_{\theta}\ \mathbb{E}_{(\tau,l)\sim\mathcal{D}}
\left[
\sum_{t=1}^{T} \ell\big(\pi_{\theta}(o_{\leq t}, l), a_t\big)
\right],
\end{equation}
where $\ell(\cdot,\cdot)$ denotes the task-specific action supervision objective.

\textbf{Vision-Language-Action Models.}
A typical VLA model consists of three primary components: a visual encoder, an LLM backbone, and an action decoder. The visual encoder extracts visual features from input observations, while the LLM backbone integrates these visual representations with the language instruction to perform multimodal reasoning and action conditioning. Finally, the action decoder maps the resulting representations to the robot action space for execution.
Existing VLA models mainly differ in how actions are represented and learned. For instance, OpenVLA formulates action generation as \emph{next-token prediction}, where discretized actions are generated autoregressively. In contrast, $\pi_0$ adopts \emph{flow matching} to model continuous actions, while OpenVLA-OFT employs \emph{parallel decoding}, predicting multiple future actions simultaneously using bidirectional attention for efficient chunk-level prediction.
Despite these differences, all approaches rely on the LLM's ability to effectively interpret both visual observations and language instructions, which is critical for accurate action prediction.
In this work, we focus on investigating how visual information influences the action prediction performance of VLA models.

\begin{figure*}[t] 
    \centering
    % \vspace{-0.2cm}
    \includegraphics[width=\textwidth]{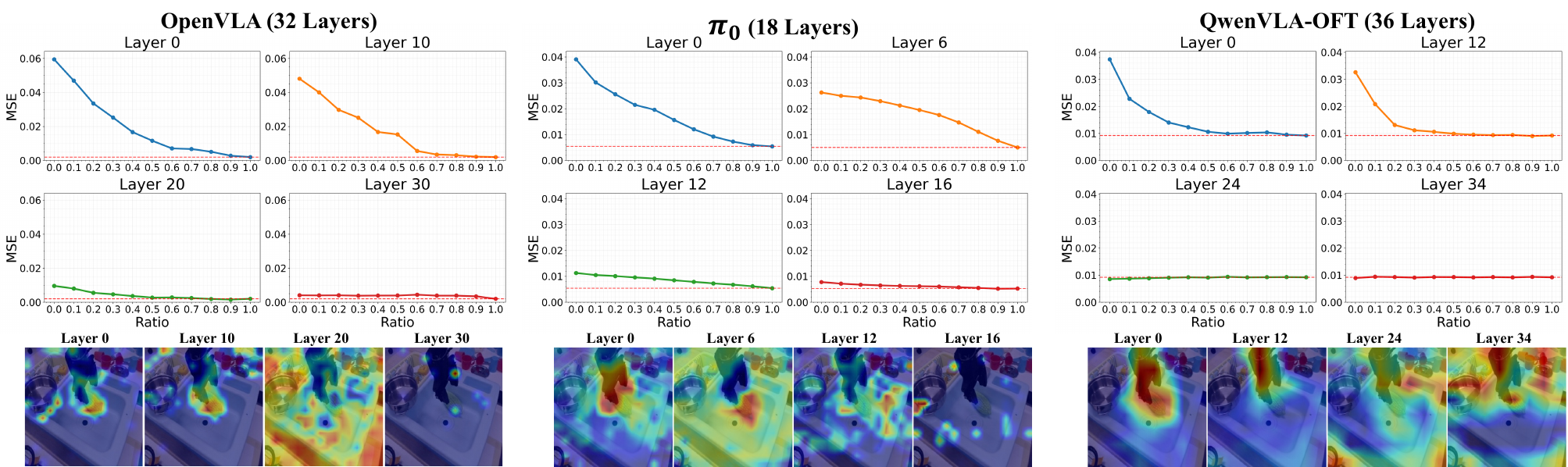} 
    \vspace{-0.5cm}
    \caption{Analysis of visual grounding in VLA models.
Top: Effect of masking ROI visual tokens on action prediction error (MSE) at different layers. Masking in shallow layers significantly degrades performance, while the impact diminishes in deeper layers.
Bottom: Visualizing the influence of visual tokens on action prediction. Attention is concentrated on task-relevant regions in shallow layers but becomes increasingly diffuse in deeper layers, indicating weakened visual grounding.}
    \label{fig:motivation}
    \vspace{-0.1cm}
\end{figure*}

\subsection{Probing the Role of Vision in VLA Models}
\label{sec:visual_drift}
Given that the LLM backbone in VLAs is composed of stacked Transformer layers, we conduct a layer-wise investigation to understand how visual information is processed throughout action prediction. 
Specifically, we first analyze action-to-vision attention maps to determine whether the model effectively attends to task-relevant objects in the scene, as such grounding provides a measure of reliable action learning. 
Next, to quantitatively evaluate the contribution of visual tokens to action performance, we perform a controlled, layer-wise ablation by masking critical visual tokens and observing the corresponding changes in action prediction accuracy.
In the following, we describe our experimental setup and present the observed results along with their analysis.

\textbf{Experimental Setup.}
We evaluate three representative VLA models that differ in their LLM backbones, model depths, and action generation paradigms: OpenVLA \cite{kim2024openvla}, $\pi_0$ \cite{black2024pi_0}, and a custom baseline QwenVLA-OFT, which adopts Qwen3-VL \cite{bai2025qwen3} as the backbone and performs parallel action prediction with an $\ell_1$ regression objective.
Our analysis is conducted on 1{,}500 randomly sampled trajectories from the BridgeV2 \cite{walke2023bridgedata} dataset, which offers high-quality manipulation demonstrations with clear object layouts and consistent visual observations, making it particularly suitable for studying object-level visual sensitivity. For each image, we employ Grounding-DINO \cite{liu2024grounding} to localize the regions of interest, including the robot arm, the manipulated object, and their interaction area.

\textbf{Layer-wise Visual Token Contribution to Action Prediction.} 
To better understand how VLA models ground action prediction in visual information, we analyze the contribution of visual tokens across LLM layers using Grad-CAM~\cite{selvaraju2017grad}. 
Specifically, for each layer, we compute gradient-based contribution scores of visual tokens with respect to the predicted action, and visualize the resulting token-wise contribution map on the image. As shown in Figure \ref{fig:motivation} (bottom), all three VLA paradigms exhibit a consistent pattern: in relatively shallow layers, high-contribution tokens are mainly concentrated on task-relevant visual regions, including the manipulated object and the robot arm. In deeper layers, however, the contribution map becomes increasingly diffuse and shifts toward less relevant regions, indicating that action prediction gradually becomes less grounded in task-relevant visual evidence along the LLM backbone.

\textbf{Action Prediction Sensitivity to ROI Visual Tokens.}
While attention visualization provides qualitative insights into how models attend to visual regions, it does not directly quantify how much action prediction actually depends on those regions. To more rigorously measure the contribution of task-relevant visual information, we perform a layer-wise masking study on visual tokens corresponding to the ROIs.
Specifically, for a selected layer in LLM, we identify the visual tokens associated with the ROIs and zero out a fraction $r$ of them, effectively removing their information from the model, while keeping all non-ROI tokens unchanged. The resulting hidden states are then propagated through the remaining layers to produce the final action prediction.
We measure the effect of this intervention using the mean squared error (MSE) between the predicted actions and the ground-truth actions. Since the masking is restricted to ROI tokens at a specific layer, the resulting change in performance directly reflects the extent to which grounded visual information contributes to action prediction at that depth.
As shown in Figure \ref{fig:motivation} (top), all three models exhibit a consistent layer-wise pattern. 
Masking ROI tokens in early layers leads to a substantial increase in action MSE, indicating that these layers rely heavily on task-relevant visual information. 
In contrast, the effect of masking progressively decreases in deeper layers, and even the complete removal of ROI tokens in the deepest layers causes only minor changes in action prediction. 
These results indicate that task-relevant visual cues become increasingly underutilized in deeper layers, which partially undermines action reliability.

\begin{figure*}[t] 
    \centering
    % \vspace{-0.2cm}
    \includegraphics[width=\textwidth]{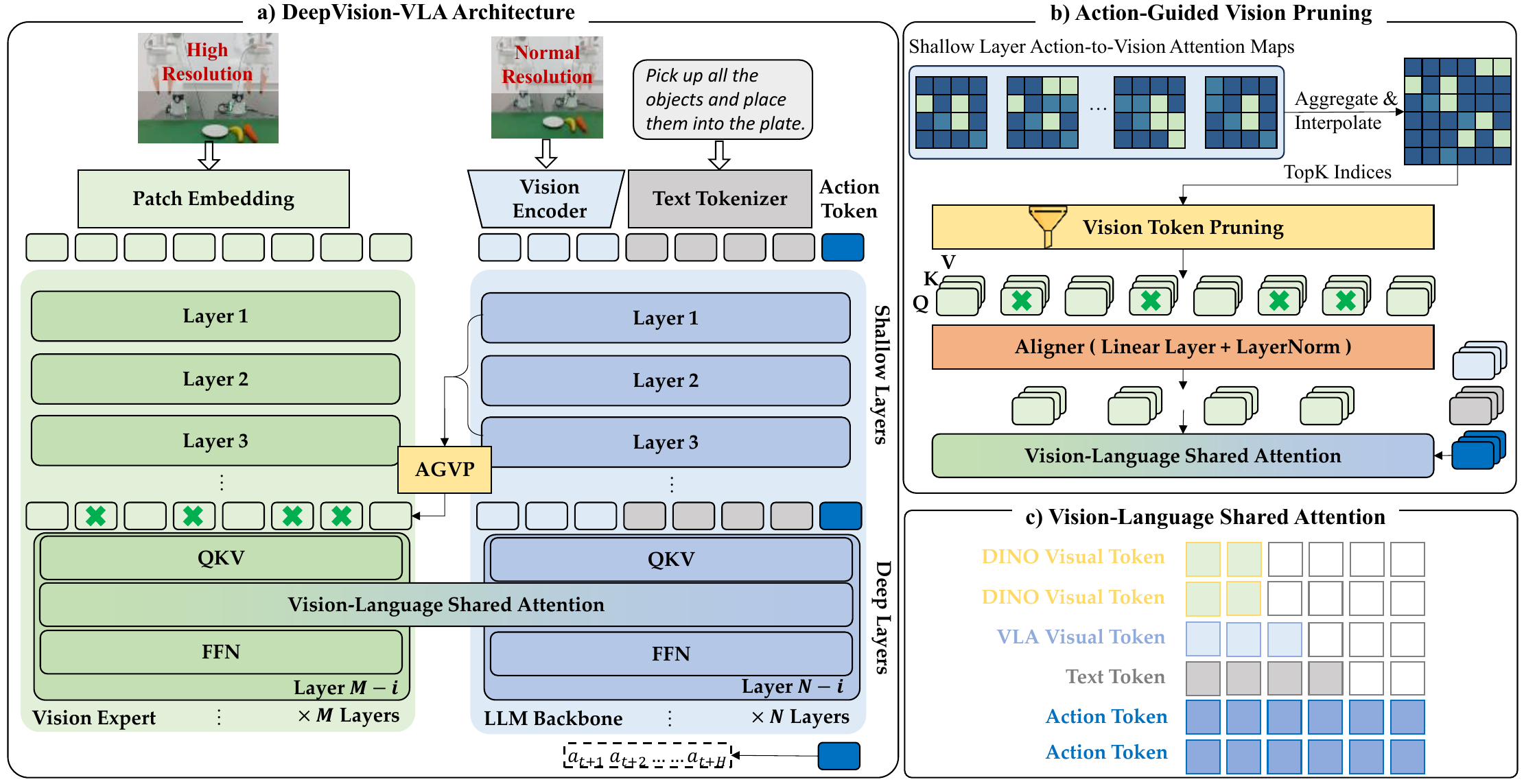} 
    \vspace{-0.5cm}
    \caption{Framework.
(a) A high-resolution Vision Expert is coupled with the LLM backbone through the proposed Vision–Language Mixture-of-Transformers (VL-MoT) framework, where deep LLM layers share attention with the Vision Expert to enhance visual grounding for action prediction.
(b) Action-to-vision attention from shallow LLM layers is aggregated to identify task-relevant regions, which are used to prune Vision Expert tokens before fusion.
(c) Vision Expert tokens use bidirectional attention to preserve their pretrained knowledge. VLA tokens apply causal attention to prompts and bidirectional attention to action tokens for parallel prediction.}
    \label{fig:method}
    % \vspace{-0.1cm}
\end{figure*}

\subsection{DeepVision-VLA}
Based on the above analysis and the insight that increasing the sensitivity of deep VLA layers to task-relevant visual ROIs may improve action prediction accuracy, we propose the Vision-Language Mixture-of-Transformers (VL-MoT) framework. 
To facilitate understanding, we illustrate how the framework is applied to the QwenVLA-OFT baseline to construct DeepVision-VLA, as shown in Figure \ref{fig:method} (a). 
% Since the proposed framework is model-agnostic, we also provide additional implementations built upon other baseline architectures in the Appendix. 
In this section, we begin by introducing the baseline model, followed by the architectural design of AVL-MoT, and finally present the Action-Guided Vision Pruning (AGVP) strategy.

\subsubsection{\textbf{QwenVLA-OFT Baseline}}
\label{sec:qwenvla-oft baseline}
The baseline model is built upon Qwen3-VL (4B) \cite{bai2025qwen3} and adopts the parallel action prediction paradigm with an $\ell_1$ regression objective as proposed in OpenVLA-OFT \cite{kim2025fine}. 
Different from OpenVLA-OFT, which applies bidirectional attention to all tokens, we restrict bidirectional attention to action tokens while keeping causal attention for prompt tokens to better preserve the pretrained behavior of Qwen3-VL.
QwenVLA-OFT consists of three main components: a visual encoder, an LLM backbone, and an action decoder.

\textbf{Visual Encoder.}  
QwenVLA-OFT utilizes SigLIP2-Large (0.3B) as its visual encoder to extract expressive image representations. The encoder employs 2D-RoPE with interpolated absolute positional embeddings to provide spatial information for visual tokens. Given an input image $I \in \mathbb{R}^{H \times W \times 3}$, it is partitioned into patches to produce dense visual features. To reduce token count and improve computational efficiency, neighboring $2 \times 2$ features are merged into a single token and projected to the LLM hidden dimension $d$ via a two-layer MLP, resulting in $\mathbf{V} = E_{\mathrm{vis}}(I) \in \mathbb{R}^{N_v \times d}$, where $N_v$ is the number of visual tokens.

\textbf{LLM Backbone.}  
The visual features $\mathbf{V}$ are concatenated with tokenized language instructions and action tokens initialized as zero vectors to form the multimodal input sequence to the LLM. At layer $\ell$, the hidden states are decomposed as $\mathbf{Z}^{\ell} = [\mathbf{Z}_V^{\ell}; \mathbf{Z}_L^{\ell}; \mathbf{Z}_A^{\ell}]$, where $\mathbf{Z}_V^{\ell}$, $\mathbf{Z}_L^{\ell}$, and $\mathbf{Z}_A^{\ell}$ denote the visual, language, and action embeddings, respectively. The LLM updates these hidden states layer by layer according to $\mathbf{Z}^{\ell} = \mathcal{F}^{\ell}(\mathbf{Z}^{\ell-1})$ for $\ell=1,\dots,L$, and the final action embedding $\mathbf{Z}_A^{L}$ is then fed to the action decoder.

\textbf{Action Decoder.}  
The action decoder maps the final action embeddings $\mathbf{Z}_A^{L}$ to the robot's action space using a two-layer MLP. Specifically, the predicted action vector $\mathbf{a} \in \mathbb{R}^{n}$, where $n$ is the number of degrees of freedom (DoF) of the embodiment.

\subsubsection{\textbf{VL-MoT Framework}}
To enhance visual grounding in the deeper layers of the VLA, we introduce a \textbf{Vision Expert}, DINOv3. Its multi-level transformer features are injected into deep VLA layers via a shared-attention mechanism. This design allows the VLA to incorporate multi-level visual information into layers that would otherwise be insensitive to task-relevant visual ROIs, thereby improving the precision of action generation.

\textbf{Vision Expert.} 
We adopt the visual foundation model DINOv3 \cite{simeoni2025dinov3} because it provides spatially detailed representations \cite{kim2024openvla,li2024cogact,kim2025fine} that are more fine-grained than the high-level semantic features typically extracted by existing VLA visual encoders, making it particularly suitable for precise manipulation tasks. Our approach, however, is not limited to this choice, and exploring alternative vision experts is left for future work.

\textbf{VL-MoT Design.} 
To integrate the multi-level knowledge of the Vision Expert into the deep VLA layers, we do not simply concatenate intermediate features and process them jointly through the LLM.  
Instead, we introduce a novel MoT architecture that directly exposes the intermediate Query, Key, and Value (QKV) representations from the Vision Expert and integrates them with the corresponding QKV of the deep VLA layers via a shared-attention mechanism, as shown in Figure \ref{fig:method} (a).  
This enables cross-branch information exchange while preserving separate processing pathways, effectively reducing feature interference and stabilizing fusion training.

To align with the layers where visual grounding becomes less sensitive, the Vision Expert is connected only to the deepest $n$ layers of the VLA, where $n$ is determined based on the analysis in Sec.~3.2. 
For the Vision Expert, we select the last $n$ transformer layers and use their QKV representations for feature injection. 
To formally describe this interaction, we consider a coupled layer between the Vision Expert and the LLM backbone.  
Here, $\mathbf{E}^{k} \in \mathbb{R}^{M \times d_e}$ represents the features from the $k$-th transformer layer of the Vision Expert, where $M$ is the number of expert tokens and $d_e$ is the expert hidden dimension.  
Similarly, $\mathbf{Z}^{\ell} \in \mathbb{R}^{N \times d}$ denotes the token representations of the LLM at layer $\ell$, where $N$ is the number of LLM tokens and $d$ is the LLM hidden dimension.
We then compute the QKV projections for the Vision Expert and the LLM tokens separately:
\begin{equation}
\begin{aligned}
Q_E = \mathbf{E}^{k} W_Q^{E}, \quad
K_E = \mathbf{E}^{k} W_K^{E}, \quad
V_E = \mathbf{E}^{k} W_V^{E}, \\
Q_Z = \mathbf{Z}^{\ell} W_Q^{Z}, \quad
K_Z = \mathbf{Z}^{\ell} W_K^{Z}, \quad
V_Z = \mathbf{Z}^{\ell} W_V^{Z}.
\end{aligned}
\end{equation}
Since the hidden dimensions of the two branches may differ, the QKV representations of the Vision Expert tokens are each projected through a learnable linear layer to align with the LLM hidden dimension. The shared attention is then computed by concatenating the QKV representations from both branches:
\begin{equation}
Q = [Q_E ; Q_Z], \quad 
K = [K_E ; K_Z], \quad 
V = [V_E ; V_Z].
\end{equation}
\begin{equation}
A = \mathrm{softmax}\!\left(\frac{Q K^\top}{\sqrt{d_k}}\right), \quad
H = A V.
\end{equation}
Finally, we split the output $H = [H_E ; H_Z]$ back into the two branches and apply the remaining standard Transformer operations. Notably, we maintain the original bidirectional attention mechanism for visual expert tokens, as shown in Figure \ref{fig:method} (c), which better preserves pre-trained knowledge and provides more stable visual features for the VLA model.

Empirically, this VL-MoT configuration outperforms alternative feature selection strategies, such as using QKV features exclusively from the first $n$ layers or uniformly sampling $n$ layers across the entire Vision Expert.
We attribute this design choice to the fact that the deeper layers of the Vision Expert encode high-level, semantically rich, and object-centric representations, which align naturally with the task-relevant, action-conditioned features in the VLA model and are particularly effective for fine-grained and complex manipulation.

\subsubsection{\textbf{Action-Guided Vision Pruning}}
To enable the VLA model to better focus on task-relevant objects, we propose Action-Guided Visual Pruning, as shown in Figure \ref{fig:method} (b), which filters redundant background features before integrating multi-level information from the Vision Expert into deeper VLA layers.

Based on our observation in Section~3.2 that shallow VLA layers provide reliable action-conditioned visual grounding, we use the attention from action tokens to visual tokens in these layers to identify ROI visual tokens. Let $\mathbf{A}^{\ell} \in \mathbb{R}^{N_a \times N_v}$ denote the action-to-vision attention map at layer $\ell$, where $N_a$ and $N_v$ are the numbers of action and visual tokens, respectively. 
% Since multiple action tokens jointly contribute to robot control, we first average their attention as 
% $\mathbf{m}^{\ell} = \frac{1}{N_a} \sum_{i=1}^{N_a} \mathbf{A}^{\ell}_{i,:}$ to obtain a per-layer attention map. 
% We then select the layer whose attention map best focuses on task-relevant objects and use it as the indicator for subsequent pruning. In practice, this selection can be viewed as a hyperparameter search over candidate layers. This procedure effectively filters out noisy activations and yields stable, task-relevant ROI visual tokens.
Since multiple action tokens jointly contribute to robot control, we first average over the action tokens as $\mathbf{m}^{\ell} = \frac{1}{N_a} \sum_{i=1}^{N_a} \mathbf{A}^{\ell}_{i,:}$ to obtain a per-layer attention map, and then aggregate over a set of shallow layers $\mathcal{L}_s$ to obtain the final attention map $\mathbf{m} = \frac{1}{|\mathcal{L}_s|} \sum_{\ell \in \mathcal{L}_s} \mathbf{m}^{\ell}$.
This multi-layer averaging effectively filters out noise and produces stable task-relevant ROI visual tokens.

Compared with the input image used in the VLA, we leverage the strong feature extraction capability of the Vision Expert by feeding a higher-resolution image, allowing the model to capture more fine-grained object details. Consequently, we interpolate the obtained attention map to match the resolution of the Vision Expert. Denoting the interpolation operator by $\mathcal{I}(\cdot)$, the transferred attention map is $\tilde{\mathbf{m}} = \mathcal{I}(\mathbf{m}) \in \mathbb{R}^{N_d}$, where $N_d$ is the number of Vision Expert image tokens. Since this attention map effectively indicates the importance of each region, we retain the top-$K$ tokens with the highest importance, whose indices are given by $\mathcal{S}_K = \operatorname{TopK}(\tilde{\mathbf{m}}, K)$, and the corresponding tokens for sharing attention with deep VLA layers are $\bar{\mathbf{E}}^k = \mathbf{E}^k[\mathcal{S}_K]$. This approach not only effectively injects critical visual region information into deep VLA layers but also controls the computational budget while exploiting the high-resolution input of the Vision Expert.

\subsection{Training and Inference}
\textbf{Training Recipe.}
Before pretraining DeepVision-VLA, we initialize the model with pretrained weights from Qwen3-VL \cite{bai2025qwen3} and DINOv3 \cite{simeoni2025dinov3}, and then train the entire architecture in an end-to-end manner. Following prior work \cite{liu2025hybridvla,liu2026last,chen2025fast,gu2025manualvlaunifiedvlamodel}, we curate a specialized pretraining dataset by carefully processing and filtering several large-scale cross-embodiment datasets, including Open X-Embodiment \cite{o2024open}, DROID \cite{khazatsky2024droid}, and RoboMIND \cite{wu2024robomind}. The resulting dataset contains over 400K trajectories, and DeepVision-VLA is trained on this dataset for one epoch. 
The fine-tuning settings for downstream tasks are described in the experimental section, as they differ between simulation and real-world environments.

\textbf{Inference.}
At inference time, DeepVision-VLA first takes the current image observation and language instruction as input. These inputs are propagated through the shallow layers of the VLA, where action-to-vision attention maps are computed to identify task-relevant visual regions. Based on these cues, we prune the multi-level features from the Vision Expert and integrate them into the deeper VLA layers through the VL-MoT mechanism. Finally, after the deep layers complete the forward pass, the hidden states of the action tokens are fed into the action decoder to generate the final actions. Notably, this procedure for identifying key visual regions and enhancing VLA perception requires no additional annotations or external supervision, and the entire pipeline remains end-to-end executable after training.

% \clearpage

\section{Experiments}

\begin{table*}[t]
% \vspace{-0.1cm}
\caption{
\textbf{Comparison of DeepVision-VLA and baselines on RLBench.} All methods are trained in the multi-task setting~\cite{shridhar2022peract}, and we report mean success rates (S.R.).
% Inference speed is evaluated on an NVIDIA 4090 GPU.
}
\vspace{-0.1cm}
\centering
\small
\resizebox{\textwidth}{!}{
\begin{tabular}{lcccccccccc|c}
\toprule
\multirow{2}{*}{Models} & Close & Close & Toilet & Sweep & Close & Phone & Umbrella & Frame & Wine at & Water & Mean   \\
& box & laptop lid & seat down & to dustpan & fridge & on base & out & off hanger & rack & plants &   S.R. $\uparrow$ \\
\midrule
OpenVLA  & 0.60 & 0.35 & 0.75 & 0.55 & 0.85 & 0.20 & 0.30 & 0.15 & 0.20 & 0.05  & 0.40  \\
SpatialVLA & 0.80 & 0.70 & 0.85 & 0.20 & 0.80 & 0.15 & 0.25 & 0.40 & 0.15 & 0.30 & 0.46  \\
CogACT & 0.90 & 0.80 & 0.95 & 0.50 & 0.85 & 0.50 & 0.55 & 0.45 & 0.30 & 0.25 & 0.61   \\
CoT-VLA & 0.95 & 0.75 & \textbf{1.00} & 0.80 & 0.65 & 0.50 & 0.40 & 0.50 & 0.55 & \textbf{0.50} & 0.66   \\
$\pi_{0.5}$ & 0.90 & \textbf{0.95} & 0.85 & 0.75 & \textbf{1.00} & 0.05 & 0.10 & \textbf{0.80} & 0.75 & 0.35 & 0.65  \\
HybridVLA  & 0.85 & \textbf{0.95} & \textbf{1.00} & 0.90 & \textbf{1.00} & 0.50 & 0.50 & 0.70 & 0.50 & \textbf{0.50} & 0.74  \\ 
QwenVLA-OFT & 0.95 & \textbf{0.95} & \textbf{1.00} & 0.15 & 0.95 & 0.65 & 0.30 & 0.70 & 0.65 & \textbf{0.50} & 0.69\\
DeepVision-VLA  & \textbf{1.00} & \textbf{0.95} & \textbf{1.00} & \textbf{0.95} & 0.95 & \textbf{0.75} & \textbf{0.65} & 0.70 & \textbf{0.85} & \textbf{0.50} & \textbf{0.83}   \\
\bottomrule
\end{tabular}}
\vspace{-0.3cm}
\label{tab:rlbench}
\end{table*}

In Section~\ref{sec: SE}, we systematically evaluate the manipulation capability of our proposed DeepVision-VLA in simulated environments. Subsequently, Section~\ref{sec: AS} presents an ablation study to validate the essential roles of the VL-MoT framework and AGVP strategy. Finally, in Section~\ref{sec: RE}, we further demonstrate the model's effectiveness through single-arm real-world manipulation tasks.

\subsection{Simulation Experiment}
\label{sec: SE}

\textbf{Data Collection.} 
We evaluate DeepVision-VLA across 10 manipulation tasks from the RLBench~\cite{james2020rlbench} benchmark based on the CoppeliaSim simulation environment, including 1) \textit{Close box}, 2) \textit{Close Laptop}, 3) \textit{Toilet seat down}, 4) \textit{Sweep to dustpan}, 5) \textit{Close fridge}, 6) \textit{Phone on base}, 7) \textit{Take umbrella out}, 8) \textit{Frame off hanger}, 9) \textit{Wine at rack}, and 10) \textit{Water plants}. All tasks are performed on a Franka Panda robot with a single front-view RGB observation. Following pre-defined waypoints and utilizing the Open Motion Planning Library~\cite{sucan2012open}, we construct a training dataset in which each task comprises 100 trajectories, with the frame-sampling technique used in previous studies~\cite{shridhar2022peract}. 

\textbf{Training and evaluation protocol.} 
We compare DeepVision-VLA against seven representative VLA baselines: OpenVLA~\cite{kim2024openvla}, SpatialVLA~\cite{qu2025spatialvla}, CogACT~\cite{li2024cogact}, CoT-VLA~\cite{zhao2025cot}, $\pi_{0.5}$~\cite{intelligence2025pi_}, HybridVLA~\cite{liu2025hybridvla}, and our custom QwenVLA-OFT. Each baseline is initialized with its officially released pretrained weights and fine-tuned end-to-end, strictly following the configurations and hyperparameters recommended in their original publications, except for QwenVLA-OFT, which we pretrain ourselves. The implementation details of QwenVLA-OFT are described in Section~\ref{sec:qwenvla-oft baseline}.
For image processing, DeepVision-VLA adopts dual resolutions: 256×256 for the VLA branch and 512×512 for the Vision Expert, providing finer visual details to the Vision Expert branch. We integrate the last 16 layers of DINOv3-H into the last 16 layers of Qwen3VLA-OFT, set the pruning ratio to 0.5. We fine-tune DeepVision-VLA for 300 epochs using the AdamW~\cite{loshchilov2017decoupled} optimizer on 8 NVIDIA H20 GPUs.
Following standard evaluation protocols in prior works~\cite{gu2025manualvlaunifiedvlamodel, goyal2023rvt}, we report results from the final checkpoints across 20 rollout trials per task, repeated over three random seeds to account for stochastic variability.

\textbf{Quantitative analysis.} 
As shown in Table~\ref{tab:rlbench}, DeepVision-VLA achieves an 83\% mean success rate on RLBench, outperforming all baselines by a significant margin. It surpasses HybridVLA (74\%), $\pi_{0.5}$ (65\%), and CogACT (61\%) by 9\%, 18\%, and 22\%, respectively, and improves over our direct baseline QwenVLA-OFT (69\%) by 14\%, directly validating the effectiveness of our proposed Vision-Language Mixture-of-Transformers (VL-MoT) framework coupled with the Action-Guided Visual Pruning (AGVP) strategy.
DeepVision-VLA attains the highest success rate on 8 out of 10 tasks, demonstrating strong robustness across diverse manipulation scenarios, with particularly large gains on visually challenging tasks. For instance, on \textit{Sweep to Dustpan} and \textit{Wine at Rack}, it improves over QwenVLA-OFT by 80\% (0.95 vs. 0.15) and 31\% (0.85 vs. 0.65), respectively.
These improvements stem from its enhanced visual processing: by injecting multi-level DINOv3 features into deeper layers and pruning irrelevant visual tokens, the model maintains strong task-relevant visual grounding throughout the model, addressing a key limitation of standard VLAs and yielding substantial performance gains.

\subsection{Ablation Study}
\label{sec: AS}

To validate the key design choices of DeepVision-VLA, we conduct ablation experiments on four representative RLBench tasks: 1) \emph{Close box}, 2) \emph{Close laptop}, 3) \emph{Sweep to dustpan}, and 4) \emph{Phone on base}. These tasks are selected to cover both relatively simple and more challenging manipulation scenarios, while remaining representative and discriminative for evaluating visual grounding quality. All models are trained under the same setting and evaluated over 50 rollouts per task. We report the average success rate over the four tasks.

\textbf{Importance of VL-MoT Architecture.} 
We first study how different paradigms for incorporating Vision Expert features affect performance. Across the four evaluation tasks, the vanilla VLA baseline (QwenVLA-OFT) achieves a 65.5\% success rate. Early fusion of DINOv3 features with the original visual tokens before projection~\cite{kim2024openvla} improves performance to 73\%, showing the benefit of external vision features, but its shallow input-level integration remains limited. Aligning intermediate VLA features to frozen DINOv3 representations~\cite{kachaev2025don} yields 67\%, suggesting that preserving visual representations helps, though the aligned features are still generic rather than manipulation-specific. In contrast, our proposed VL-MoT framework achieves 88\%, substantially outperforming all alternatives. This demonstrates that directly coupling the Vision Expert with deeper VLA layers, together with targeted feature pruning, is a more effective way to enhance action-relevant visual representations.

\begin{figure*}[t] 
    \centering
    % \vspace{-0.2cm}
    \includegraphics[width=\textwidth]{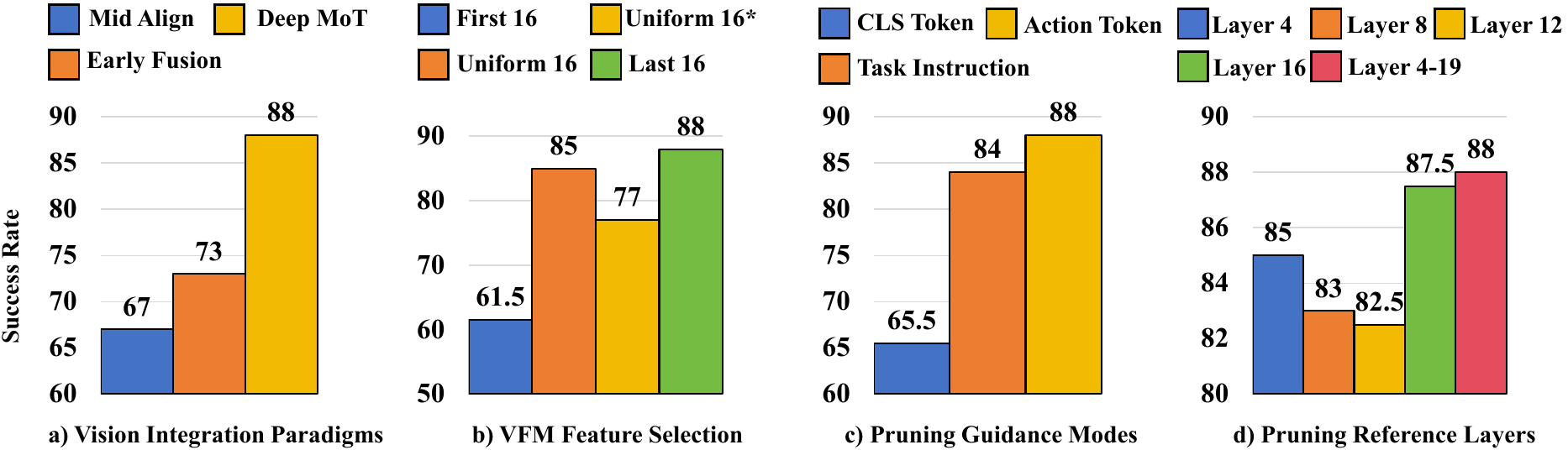} 
    % \vspace{-0.7cm}
    \caption{Ablation studies.
(a) Comparison of different paradigms for integrating Vision Expert features into the VLA backbone.
(b) Evaluation of different multi-level feature selection strategies from the Vision Expert. `Uniform 16*' signifies the use of uniformly sampled 16-layer features from the VLA's SigLIP encoder.
(c) Comparison of different guidance mechanisms for generating the visual pruning map in AGVP.
(d) Analysis of different shallow-layer references used to compute the action-to-vision attention map for pruning.
}
    \label{fig:ablation}
    \vspace{-0.5cm}
\end{figure*}
\textbf{Vision Foundation Model Feature Selection.} 
We next examine both the feature selection strategy and the choice of the Vision Foundation Model. Using only the first 16 layers of DINOv3 yields 61.5\%, while uniformly sampling 16 layers across the whole model achieves 85\%. The best result is obtained by using the last 16 layers, which reaches 88\%, indicating that later DINOv3 features are the most effective for robotic manipulation. Replacing DINOv3 with SigLIP under the same uniformly sampled 16-layer setting reduces the success rate to 77\%. This gap likely stems from their different pretraining objectives: SigLIP emphasizes image-text alignment, whereas DINOv3 provides stronger fine-grained spatial representations that are better suited for precise manipulation.

\textbf{Guidance Mechanisms for Action-Guided Vision Pruning.} 
We then evaluate different strategies for generating the guidance map used in AGVP. The baseline achieves 65.5\%. Using the DINOv3 CLS token as the pruning cue brings no improvement, remaining at 65.5\%, suggesting that global scene semantics alone are insufficient for manipulation-oriented feature selection. Using instruction-to-vision attention improves the success rate to 84\%, showing that task-aware language guidance is beneficial, but its lack of reliable awareness of the robot arm and the arm-object interaction region limits its effectiveness for precise action prediction. Our action-to-vision attention guidance achieves the best result of 88\%, indicating that shallow-layer action tokens provide more effective cues for identifying manipulation-relevant regions, as they naturally encode both task intent and action-conditioned visual grounding.

\textbf{Impact of Reference Layers on AGVP.}
We further analyze which shallow-layer attention maps provide the most effective pruning guidance. Using the attention map from the 4th, 8th, 12th, and 16th layers yields success rates of 85\%, 69\%, 82.5\%, and 87.5\%, respectively, with the 16th layer performing best among single-layer choices. Averaging attention maps across Layers 4-19 further improves performance to 88\%, indicating that multi-layer guidance is more robust and better suppresses noisy attention to irrelevant regions.

% 需在导言区引用: \usepackage{booktabs, graphicx}

\begin{table*}[t]
\centering
\caption{\textbf{Comparison across real-world manipulation tasks. Step represents the atomic task within the overall task, and Avg denotes the average success rate. Our method is built upon the QwenOFT-VLA baseline.}}
\vspace{-2mm}
\small
\setlength{\tabcolsep}{4pt} 
\renewcommand{\arraystretch}{1.3}
\resizebox{0.99\textwidth}{!}{
\begin{tabular}{p{3.0cm} cc cc cc c} 
\toprule
\multirow{2}{*}{\textbf{Models}} & \multirow{2}{*}{\textbf{Stack coke cans}} & \multirow{2}{*}{\textbf{Write letter `S'}} & \multicolumn{2}{c}{\textbf{Pick fruit to the plate}} & \multicolumn{2}{c}{\textbf{Pour coke to bottle}} & \multirow{2}{*}{\textbf{Avg}} \\
\cmidrule(lr){4-5} \cmidrule(lr){6-7} 
& & & Step 1 \hspace{0.2cm}$\rightarrow$ & \hspace{-0.25cm}Step 2 & Step 1 \hspace{0.2cm}$\rightarrow$ & \hspace{-0.10cm}Step 2 & \\
\midrule
$\pi_{0.5}$ & 0.65 & 0.95 & 0.75 & 1.00 & 1.00 & 0.70 & 0.842 \\
OpenVLA-OFT & 0.50 & 0.85 & 0.70 & 0.80 & 0.75 & 0.70 & 0.717 \\
QwenVLA-OFT & 0.50 & 0.80 & 0.85 & 0.90 & 0.70 & 0.70 & 0.742 \\
% \midrule
% 直接在 p 列换行，右侧数字将与第一行文字对齐（顶端对齐）
\textbf{DeepVision-VLA} & \textbf{0.65} & \textbf{0.95} & \textbf{0.95} & \textbf{0.95} & \textbf{1.00} & \textbf{1.00} & \textbf{0.917} \\
\bottomrule
\end{tabular}}
\label{tab:real_world_exp}
\end{table*}

\begin{figure*}[t] 
    \centering
    % \vspace{-0.2cm}
    \includegraphics[width=\textwidth]{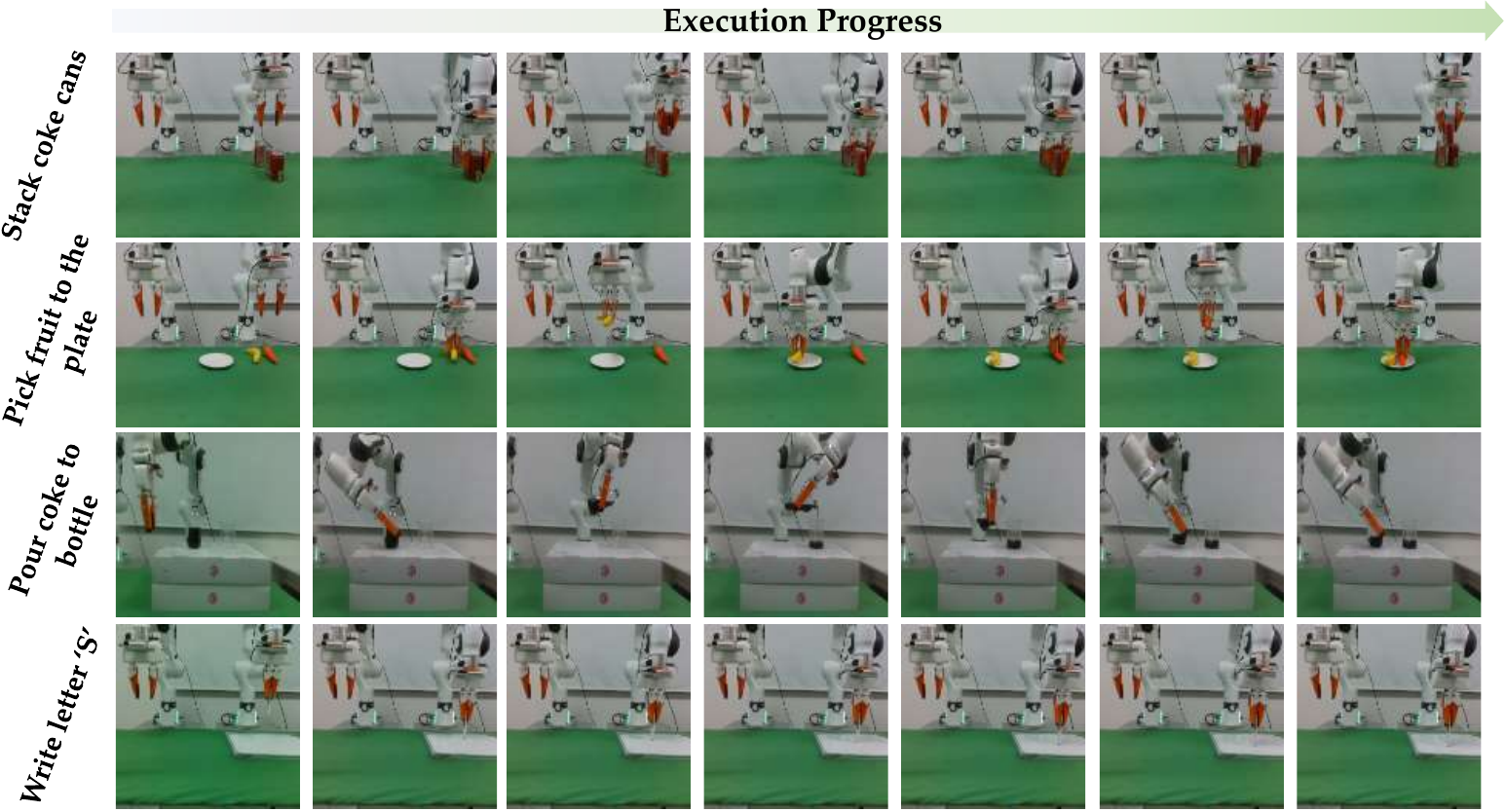} 
    % \vspace{-0.7cm}
    \caption{Visualization of task execution processes by real-world single-arm robots (left to right).
}
    \label{fig:real-world}
    \vspace{-0.3cm}
\end{figure*}

\subsection{Real-World Experiment}
\label{sec: RE}

\textbf{Data Collection.}
We evaluate DeepVision-VLA on four complex single-arm real-world manipulation tasks using a Franka Research 3 robot equipped with an Intel RealSense D455 camera for RGB observations. The evaluated tasks encompass 1) \textit{stack coke cans}, 2) \textit{write letter `S'}, 3) \textit{pick fruit to the plate}, and 4) \textit{pour coke to bottle}.
To provide a more comprehensive assessment of model performance, Tasks~3s) and 4) are decomposed into two sequential steps. Task~3) is divided into a first step of picking and placing the banana, followed by a second step of picking and placing the carrot. Task~4) is evaluated independently on the initial grasp and the subsequent pouring action. Tasks~1) and 2) are defined as simple tasks and are assessed using a single comprehensive success score for the complete execution.

\textbf{Training and evaluation details.}
The training protocol for the DeepVision-VLA remains the same as in the simulation. 
We compare our method against three state-of-the-art VLA model baselines including $\pi_{0.5}$~\cite{black2024pi_0}, OpenVLA-OFT~\cite{kim2025fine}, and the custom QwenVLA-OFT~\cite{starvla2025}. 
For a fair evaluation, all baselines share the identical camera setup, with each task undergoing 20 rollouts under consistent test conditions.

\textbf{Quantitative and qualitative analysis.}
Tab. \ref{tab:real_world_exp} shows that DeepVision-VLA achieves the highest overall performance across all real-world evaluations with an average success rate of 91.7\%, effectively surpassing the strong baselines including $\pi_{0.5}$ (84.2\%), Qwen-OFT (74.2\%), and OpenVLA-OFT (71.7\%), which demonstrates the highly robust physical interaction and precise manipulation capabilities of our DeepVision-VLA.
For the two single-step tasks evaluated with a comprehensive score, DeepVision-VLA attains success rates of 65\% for stacking coke cans and 95\% for writing the letter `S'. The whiteboard writing task particularly highlights the strengths of our architecture, as the robot must precisely draw instructed shapes requiring continuous, high-precision spatial tracking.
Furthermore, DeepVision-VLA excels in multi-stage tasks evaluated via granular sub-stage metrics. During the ``pick fruit to the plate'' task, it maintained a consistent 95\% success rate across both Step 1 and Step 2. More notably, in the ``pour coke to bottle'' task, our model achieved a perfect 100\% success rate across both stages, outperforming the strongest baseline ($\pi_{0.5}$), which dropped to 70\% in the second step. These multi-stage tasks necessitate strict visual focus on object boundaries and relative positioning.
We attribute these improvements to our VL-MoT framework and AGVP strategy, which facilitate the integration of hierarchical visual features from the Vision Expert into the deeper layers of the VLA. This design enables the model to attend to critical object regions, leading 
% \vspace{-0.5cm}
\begin{table}[t]
\centering
\caption{\textbf{Generalization Scenarios.} The visualizations show unseen test conditions, where \textbf{Background} and \textbf{Lighting} denote novel environmental layouts and varied illumination conditions, respectively. \textbf{DeepVision-VLA} exhibits robust visual enhancement, maintaining precise manipulation under these perturbations.}
\label{tab:robustness_pick_fruit}
\vspace{0.1cm}

% --- 左侧部分：图片 ---
\begin{minipage}{0.34\linewidth}
    \centering
    \includegraphics[width=\linewidth]{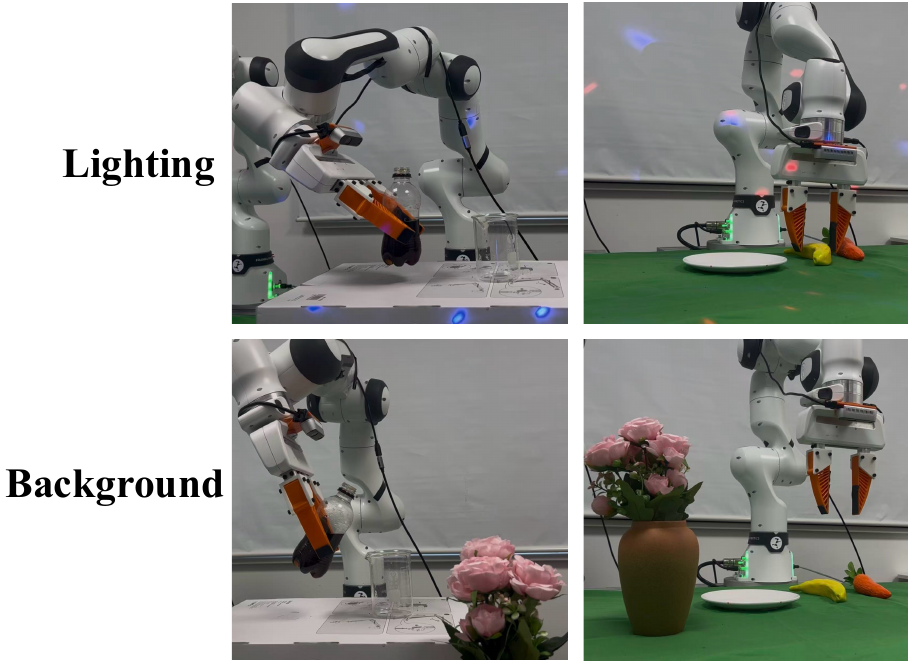}
\end{minipage}
\hfill % 填充中间空白
% --- 右侧部分：表格 ---
\begin{minipage}{0.65\linewidth}
    \centering
    \renewcommand{\arraystretch}{1.3}
    \setlength{\tabcolsep}{2pt}
    \resizebox{\linewidth}{!}{
    \begin{tabular}{l cc cc}
    \toprule
    \textbf{Task} & \multicolumn{2}{c}{\textbf{Pick fruit (Step 1)}} & \multicolumn{2}{c}{\textbf{Pick fruit (Step 2)}} \\
    \cmidrule(lr){2-3} \cmidrule(lr){4-5}
    \textbf{Scenario} & QwenVLA-OFT & \textbf{Ours} & QwenVLA-OFT & \textbf{Ours} \\
    \midrule
    Original   & 0.85 & \textbf{0.95} & 0.90 & \textbf{0.95} \\
    \midrule
    Background & 0.70 (-18\%) & \textbf{0.90 (-5\%)} & 0.80 (-11\%) & \textbf{0.90 (-5\%)} \\
    \addlinespace
    Lighting   & 0.70 (-18\%) & \textbf{0.80 (-16\%)} & 0.70 (-22\%) & \textbf{0.90 (-5\%)} \\
    \bottomrule
    \end{tabular}}
\end{minipage}
\vspace{-0.5cm}
\end{table}

to more accurate action generation. 
Figure~\ref{fig:real-world} visualizes the execution process of DeepVision-VLA on real-world tasks. The smooth manipulation and successful task completion demonstrate that our model effectively attends to task-critical visual objects. For instance, the 100\% success rate on the ``pour Coke to Bottle'' task highlights the model's robust object localization and execution stability.

\subsection{Generalization Experiment}
To evaluate the zero-shot generalization capacity of DeepVision-VLA, we conducted a series of rigorous experiments across two distinct environmental perturbations, comparing our model against the baseline QwenVLA-OFT. We focused on two representative tasks: the multi-step Pick fruit and the multi-step Pour coke.

\textbf{Unseen background}
By introducing unseen elements (e.g., floral arrangements) into the workspace, we challenged the models' ability to distinguish task-relevant features from environmental noise. As reported in the "Background" row of Table \ref{tab:robustness_pick_fruit}, while the baseline experienced a noticeable performance decay, DeepVision-VLA maintained high success rates (e.g., 0.90 in Pick fruit Step 1). This resilience stems from our visual enhancement mechanism, which enables the model to perform high-level scene reasoning and decouple the manipulated object from its surroundings.

\textbf{Unseen lighting conditions}
The results under "Lighting" further validate our model's superiority in visual robustness. While baseline methods often struggle with altered shadow patterns and non-uniform illumination, DeepVision-VLA maintains near-perfect success rates (e.g., 1.00 in Pour coke). This suggests that our visual enhancement strategy effectively achieves a degree of photometric invariance, allowing the model to extract consistent structural information even when the raw pixel intensity fluctuates significantly.

\section{Conclusion}
In this work, we investigate the role of visual representations in Vision-Language-Action (VLA) models for robotic manipulation. Through a systematic analysis across multiple VLA architectures and action-generation paradigms, we observe that sensitivity to visual tokens progressively diminishes in deeper layers during action generation, limiting the effective utilization of visual information. Motivated by this finding, we propose DeepVision-VLA, a framework built upon a Vision-Language Mixture-of-Transformers (VL-MoT) architecture that introduces shared attention between a vision foundation model and the VLA backbone. By injecting multi-level visual features into deeper layers, DeepVision-VLA enhances visual representations and improves the grounding of perception in action generation.
To further strengthen task-relevant visual cues, we introduce Action-Guided Visual Pruning (AGVP), which selectively removes irrelevant visual tokens based on shallow-layer attention while preserving critical visual information with minimal computational overhead. Extensive experiments on both simulated and real-world robotic manipulation tasks demonstrate that DeepVision-VLA consistently outperforms prior state-of-the-art methods.
We hope this work provides new insights into the design of visually enhanced VLA models and inspires future research on integrating foundation vision models with robotic manipulation.

\bibliographystyle{unsrt}
\bibliography{reference}

\end{document}